\documentclass[journal,twoside,web]{ieeecolor}
\usepackage{generic}
\usepackage{cite}
\usepackage{amsmath,amssymb,amsfonts}
\usepackage{algorithmic}
\usepackage{graphicx}
\usepackage{algorithm,algorithmic}
\usepackage{threeparttable}
\usepackage{makecell}
\usepackage{multirow}

\usepackage{amssymb}
\usepackage{latexsym}
\usepackage{xcolor}
\usepackage{url}
\usepackage{multirow}
\usepackage{framed}

\usepackage{booktabs}

\usepackage{url}
\usepackage{xcolor}

\usepackage{hyperref}
\hypersetup{hidelinks=true}
\usepackage{textcomp}
\def\BibTeX{{\rm B\kern-.05em{\sc i\kern-.025em b}\kern-.08em
    T\kern-.1667em\lower.7ex\hbox{E}\kern-.125emX}}
\begin{document}
\title{LiMT: a multi-task liver image benchmark dataset}
\author{Zhe Liu, Kai Han, Siqi Ma, Yan Zhu, Jun Chen, Chongwen Lyu, Xinyi Qiu, Chengxuan Qian, \\Yuqing Song, Yi Liu, Liyuan Tian, Yang Ji, Yuefeng Li
\thanks{This work was supported in part by the National Natural Science Foundation of China under Grant (62276116) and Jiangsu Graduate Research Innovation Program under Grant (KY-CX23$\_$3677). (Corresponding authors: Kai Han; Yuefeng Li.)}
\thanks{Zhe Liu, Kai Han, Siqi Ma, Jun Chen, Chongwen Lyu, Xinyi Qiu, Chengxuan Qian, Yuqing Song and Yi Liu are with the School of Computer Science and Communication Engineering, Jiangsu University, Zhenjiang 212013, China (e-mail: 1000004088@ujs.edu.cn; 2112108003@stmail.ujs.edu.cn; 2212208031@stmail.ujs.edu.cn; chenjun@ujs.edu.cn; 2212308023@stmail.ujs.edu.cn; 2212408042@stmail.ujs.edu.cn; chengxuan.qian@stmail.ujs.edu.cn; yqsong@ujs.edu.cn; ly@ujs.edu.cn).}
\thanks{Yan Zhu and Yang Ji are with the Department of Imaging, Affiliated Hospital of Jiangsu University, Zhenjiang 212001, China (e-mail: salary\_hi@126.com; s509167@163.com).}
\thanks{Liyuan Tian is with the Department of Information, Affiliated Hospital of Jiangsu University, Zhenjiang 212001, China (e-mail: tly\_js@163.com).}
\thanks{Yuefeng Li is with the Medical College of Jiangsu University, Zhenjiang, Jiangsu 212001, China (e-mail: jiangdalyf@163.com).}}

\maketitle

\begin{abstract}
Computer-aided diagnosis (CAD) technology can assist clinicians in evaluating liver lesions and intervening with treatment in time. Although CAD technology has advanced in recent years, the application scope of existing datasets remains relatively limited, typically supporting only single tasks, which has somewhat constrained the development of CAD technology. To address the above limitation, in this paper, we construct a multi-task liver dataset (LiMT) used for liver and tumor segmentation, multi-label lesion classification, and lesion detection based on arterial phase-enhanced computed tomography (CT), potentially providing an exploratory solution that is able to explore the correlation between tasks and does not need to worry about the heterogeneity between task-specific datasets during training. The dataset includes CT volumes from 150 different cases, comprising four types of liver diseases as well as normal cases. Each volume has been carefully annotated and calibrated by experienced clinicians. This public multi-task dataset may become a valuable resource for the medical imaging research community in the future. In addition, this paper not only provides relevant baseline experimental results but also reviews existing datasets and methods related to liver-related tasks. Our dataset is available at \url{https://drive.google.com/drive/folders/1l9HRK13uaOQTNShf5pwgSz3OTanWjkag?usp=sharing}.
\end{abstract}


\section{Introduction}
\label{sec:introduction}
\IEEEPARstart{H}{epatic} malignant tumors are one of the most common malignant tumors in the world, and both their morbidity and mortality are increasing \cite{mattiuzzi2019current}. Accurate segmentation, detection, and classification of liver lesions can assist diagnosis and treatment by enabling timely intervention and improving patient outcomes \cite{suzuki2017overview}. These methods support clinicians in identifying liver tumors \cite{jiang2022deep}, distinguishing malignant from benign lesions \cite{albain2009radiotherapy,bilic2023liver}, and locating lesions in imaging \cite{yan2020learning}, thereby facilitating appropriate treatment and improving outcomes. With advances in medical imaging, CT has become the preferred modality for assessing liver abnormalities \cite{alawneh2022livernet, miyayama2013identification}. In particular, enhanced CT not only visualizes lesion morphology and blood supply but also aids in differentiating benign from malignant lesions and evaluating tumor progression, thereby improving clinical decision-making. However, the growing number of cases increases clinicians’ workload and the risk of diagnostic errors \cite{ying2024multicenter}. To avoid the above problem, researchers have developed methods that can automatically segment, detect, and classify liver lesions to help clinicians detect early-stage tumors, judge disease severity, and adjuvant treatment.

Despite considerable progress in automatic liver lesion segmentation, detection, or classification in recent years, the capabilities for assisting comprehensive diagnosis and treatment are hampered by the lack of multi-task datasets. Several studies, such as those by Lu \textit{et al.} \cite{lu2017automatic} and Jin \textit{et al.} \cite{jin2023automatic}, have validated the clinical applicability of liver segmentation methods for volume estimation using the SLIVER07 \cite{heimann2009comparison} and 3D-IRCADb \cite{soler20103d} datasets. Nevertheless, these studies tend to ignore the characteristics of lesions and do not provide further diagnostic results. 

The introduction of the Liver Tumor Segmentation (LiTS) dataset \cite{bilic2023liver} enabled the joint segmentation of liver and tumors. Jin \textit{et al.} \cite{jin2023automatic} developed a multi-scale, high-resolution neural network that achieved strong segmentation performance on LiTS. Nevertheless, LiTS contains only malignant lesions and lacks classification labels, limiting its use in distinguishing benign from malignant tumors. Other studies \cite{yan2020learning, erickson2022class, huo2020harvesting} evaluated detection methods on the DeepLesion dataset \cite{yan2018deeplesion}. However, bounding-box annotations lack sufficient voxel-level detail, which limits precise visualization, particularly under conditions of low liver–tumor contrast. Other researchers \cite{karako2023improving} have explored direct classification methods for liver disease diagnosis, achieving highly accurate classification results. However, these approaches mainly focus on diagnostic outcomes without providing detailed spatial or morphological information. In addition, most existing studies rely on private datasets, limiting reproducibility and comparison. In summary, due to the lack of attributes, existing liver datasets can only assist in liver disease diagnosis and treatment of liver diseases from a local perspective, which is not comprehensive enough.

To address the limitation mentioned above, this paper provides the liver multi-task dataset LiMT with annotations of both anatomy and lesion types, aiming to develop interoperable data sharing and analytics solutions for comprehensively assisting the diagnosis and treatment of liver disease, as well as providing a scientific benchmark to evaluate existing solutions. A public dataset comprising 150 cases was established in collaboration with clinical partners from the Affiliated Hospital of Jiangsu University, covering four liver lesion types as well as normal liver examples. The LiMT dataset includes annotations for four common liver lesions, along with liver segmentation labels as well as lesion classification labels, making it a valuable resource for the medical imaging research community. Since previous part-task learning usually ignores the potential associations and interactions between different tasks, this may lead to a failure to capture the shared information and dependencies between tasks. Unlike part-task learning, LiMT demonstrates potential for exploring the correlations among multiple tasks and addressing concerns about the heterogeneity of task datasets during training. The contributions of this paper can be summarized as follows:
\begin{itemize}
	\item We introduce a publicly available multi-task (segmentation, detection, and classification) liver dataset that complements existing resources by enabling multi-task analysis within a homogeneous setting.
	\item We provide a review of existing datasets and methods related to the liver tasks.
	\item The dataset provides multi-label liver lesion annotations across four subtypes, encompassing visually similar but distinct lesions to support atypical cancer detection, as well as normal cases for healthy and diseased differentiation. Baseline results for major tasks are also included to facilitate further studies.
\end{itemize}

The remainder of this paper is organized as follows. Section \ref{sec2} reviews existing liver-related datasets and methods. Section \ref{sec3} describes the construction of the LiMT dataset. Section \ref{sec4} presents the statistical analysis and benchmark results. Section \ref{sec5} summarizes the full paper.

\begin{figure}[!t]
	\centering
	\centerline{\includegraphics[width=0.9\linewidth]{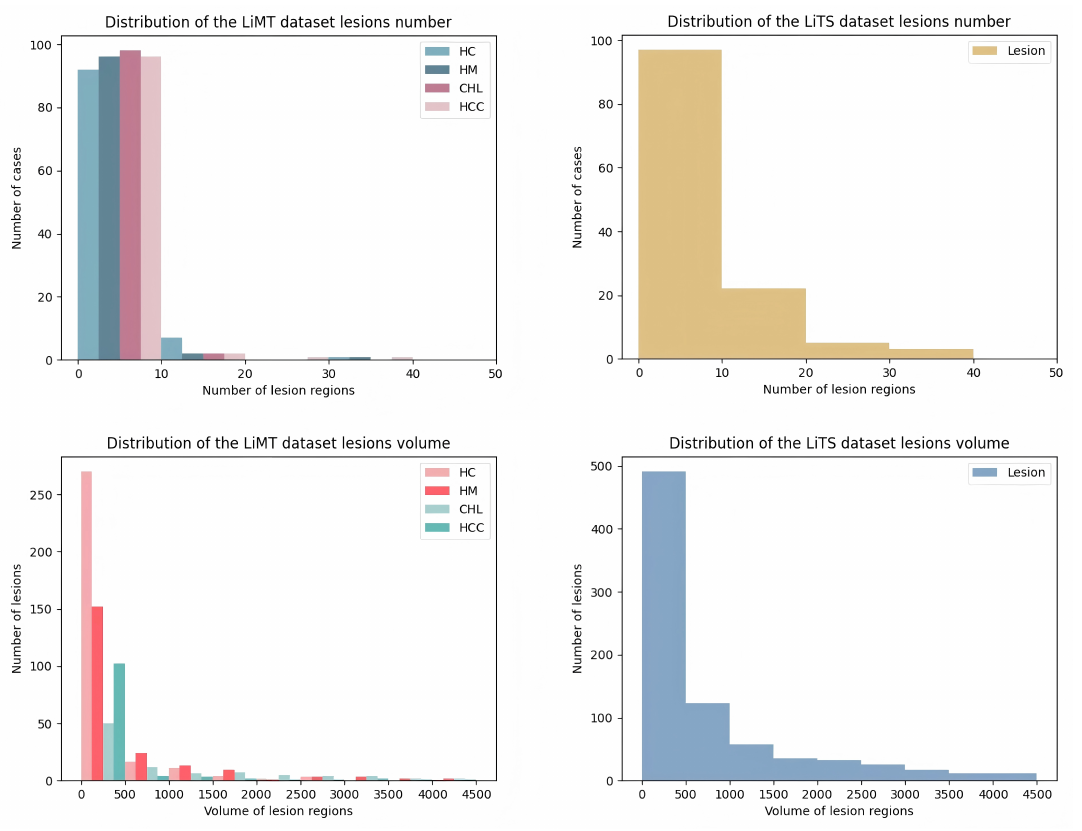}}
	\caption{The distribution of the lesion number and volume of LiTS dataset and LiMT dataset. In the figure, hepatocellular carcinoma is denoted as HCC, hepatic metastases as HM, cavernous hemangioma of liver as CHL, and liver cyst as HC.}
	\label{fig1}
\end{figure}

\section{Previous work}\label{sec2}
Section \ref{sec2.1} outlines the composition and significance of existing liver datasets. Section \ref{sec2.2} provides a detailed review of state-of-the-art methods based on these datasets, highlighting the data bottlenecks encountered by these approaches.

\subsection{Liver-related task datasets}\label{sec2.1}
This section summarizes available liver-related datasets that provide comprehensive and reliable data for disease diagnosis and treatment (see Table \ref{tab1}).

\begin{table*}[!t]
	\centering
	\caption{Datasets used for liver research}
	\label{tab1}
	\setlength{\tabcolsep}{3pt}
	\begin{threeparttable}
		\resizebox{\textwidth}{!}{
			\begin{tabular}{cccccccccc}
				\toprule
				\multirow{2}{*}{Dataset} & \multirow{2}{*}{Year} & \multicolumn{2}{c}{Modalities} & \multicolumn{3}{c}{Task} & \multicolumn{2}{c}{Num} & \multirow{2}{*}{Organ}\\
				\cline{3-9}
				&  & CT & MR & Segmentation & Detection & Classification & Train & Test & \\
				\midrule
				SLIVER07 \cite{heimann2009comparison} $\ast$ & 2007 & $\surd$ & $\times$ & $\surd$ & $\times$ & $\times$ & 20 scans & 10 scans & Liver\\
				3D-IRCADb \cite{soler20103d} $\ast$ & 2010 & $\surd$ & $\times$ & $\surd$ & $\times$ & $\times$ & \multicolumn{2}{c}{20 scans} & Liver\\
				Anatomy3 \cite{jimenez2016cloud} $\ast$ & 2015 & $\surd$ & $\surd$ & $\surd$ & $\surd$ & $\times$ & \multicolumn{2}{c}{120 scans} & 15 organs\\
				LiTS \cite{bilic2023liver} $\ast$ & 2017 & $\surd$ & $\times$ & $\surd$ & $\surd$ & $\times$ & 131 scans & 70 scans & Liver\\
				DeepLesion \cite{yan2018deeplesion} $\ast$ & 2018 & $\surd$ & $\times$ & $\times$ & $\surd$ & $\times$ & \multicolumn{2}{c}{32,120 slices} & 8 organs\\
				CT-ORG \cite{rister2020ct} $\ast$ & 2019 & $\surd$ & $\times$ & $\surd$ & $\times$ & $\times$ & 119 scans & 21 scans & \makecell{Liver / Lungs/ Bladder / \\ Kidney/ Bones / Brain}\\
				CHAOS \cite{kavur2021chaos} $\ast$ & 2019 & $\surd$ & $\surd$ & $\surd$ & $\times$ & $\times$ & 40 scans & 40 scans & Liver / Kidney / Spleen\\
				MSD \cite{simpson2019large} $\ast$ & 2019 & $\surd$ & $\surd$ & $\surd$ & $\times$ & $\times$ & \multicolumn{2}{c}{2,633 slices} & 10 organs\\
				Chen \textit{et al.} \cite{chen2019feature} $\dagger$ & 2019 & $\surd$ & $\times$ & $\times$ & $\times$ & $\surd$ & \multicolumn{2}{c}{190 scans} & Liver\\
				Naeem \textit{et al.} \cite{naeem2020machine} $\dagger$ & 2020 & $\surd$ & $\surd$ & $\times$ & $\times$ & $\surd$ & \multicolumn{2}{c}{1,200 slices} & Liver\\
				WORD \cite{luo2021word} $\ast$ & 2022 & $\surd$ & $\times$ & $\surd$ & $\surd$ & $\times$ & \multicolumn{2}{c}{150 scans} & 16 organs\\
				FLARE \cite{ma2023unleashing} $\ast$ & 2022 & $\surd$ & $\times$ & $\surd$ & $\times$ & $\times$ & \multicolumn{2}{c}{2,300 slices} & 13 organs\\
				Hussain \textit{et al.} \cite{hussain2022computer} $\dagger$ & 2022 & $\surd$ & $\times$ & $\times$ & $\times$ & $\surd$ & \multicolumn{2}{c}{1,000 slices} & Liver\\
				LLD-MMRI \cite{lou2025sdr}  $\ast$ & 2023 & $\times$ & $\surd$ & $\surd$ & $\times$ & $\times$ & \multicolumn{2}{c}{498 scans} & Liver\\
				\bottomrule
		\end{tabular}}
		\begin{tablenotes}
			\footnotesize
			\item[*] represents the public dataset.
			\item[$\dagger$] represents the private dataset.
		\end{tablenotes}
	\end{threeparttable}
\end{table*}

\subsubsection{Segmentation task}
Accurate liver segmentation is the basis for computer-assisted surgical planning \cite{qian2025adaptive,han2025climd,han2025frequency,qian2025decalign,qian2025dyncim,yuan2025video,yuan2025dvp,yuan2025autodrive,han2025region}, such as invasive surgery \cite{radtke2007computer}. As shown in Table \ref{tab2}, SLIVER07 \cite{heimann2009comparison}, released at MICCAI 2007, is a medical image dataset for evaluating liver segmentation algorithms. It comprises 20 3D CT volumes of varying sizes, resolutions, disease types, and liver morphologies, with each slice annotated by experts. \cite{jimenez2016cloud} employed cloud computing to provide a benchmarking platform for medical image segmentation, expanding the dataset to 120 cases across modalities including CT and MRI. However, algorithms are executed privately by administrators on a cloud-based virtual machine to evaluate performance on a hidden test set. This “invisible test set” lacks transparency in evaluation criteria and sample distribution, which may hinder model optimization for real-world scenarios and create a gap between benchmark performance and practical applicability.

\begin{table*}[!t]
	\centering
	\caption{Datasets used for liver segmentation task}
	\label{tab2}
	\setlength{\tabcolsep}{3pt}
	\begin{tabular}{lcccc}
		\toprule
		\multirow{2}{*}{Dataset} & Annotation & Annotation & \multirow{2}{*}{\makecell{Num (with \\ liver annotation)}} & \multirow{2}{*}{Type of lesion}\\
		& Liver & Tumor &  & \\
		\midrule
		SLIVER07 \cite{heimann2009comparison} & $\surd$ & $\times$ & 20 scans & \makecell{Metastatic hepatic adenocarcinomas (malignant), \\ Hepatic cysts (benign), Tumor (benign and malignant are not specified)}\\
		\hline
		3D-IRCADb \cite{soler20103d} & $\surd$ & $\surd$ & 20 scans & - \\
	    \hline
		Anatomy3 \cite{jimenez2016cloud} & $\surd$ & $\times$ & 120 scans & - \\
		\hline
		LiTS \cite{bilic2023liver} & $\surd$ & $\surd$ & 131 scans & \makecell{Hepatocellular carcinoma (malignant),\\ Cholangiocarcinoma (malignant), Secondary 
		liver tumors (malignant)}\\		
		\hline
		CT-ORG \cite{rister2020ct} & $\surd$ & $\times$ & 119 scans & - \\ 
		\hline
		CHAOS \cite{kavur2021chaos} &  $\surd$ & $\times$ & 20 scans & - \\
		\hline
		MSD \cite{simpson2019large} & $\surd$ & $\surd$ & 131 scans & \makecell{Hepatocellular carcinoma (malignant), Cholangiocarcinoma (malignant), \\Secondary liver  tumors (malignant)}\\
		\hline
		WORD \cite{luo2021word} & $\surd$ & $\times$ & 150 scans & - \\
		\hline
		FLARE \cite{ma2023unleashing} & $\surd$ & $\times$ & 100 scans & - \\
		\hline
		LLD-MMRI \cite{lou2025sdr} & $\times$ & $\surd$ & 498 scans &  
		\makecell{Hepatocellular carcinoma, Intrahepatic cholangiocarcinoma, Liver metastases (HM), \\Hepatic cysts (HC), Hepatic hemangioma, Focal nodular hyperplasia, and Hepatic abscess} \\
		\hline
		LiMT & $\surd$ & $\surd$ & 150 scans & \makecell{Hepatocellular carcinoma  (malignant), Hepatic  Metastases (malignant), \\ Hepatic cyst (benign),  Cavernous hemangioma of liver (benign), Normal} \\
		\bottomrule
	\end{tabular}
\end{table*}

The datasets above focus solely on liver segmentation and lack annotations for liver lesions. In clinical practice, it is essential to analyze both the liver and its lesions. Accurate segmentation of tumor lesions is crucial for cancer diagnosis, treatment planning, and monitoring treatment response \cite{bilic2023liver}. Moreover, some liver tumors do not noticeably alter liver appearance, so liver-only segmentation may miss early lesions, complicating disease assessment. To further improve the study of CAD, the 3D-IRCADb \cite{soler20103d} dataset provides images from 20 patients for both liver and tumor segmentation. However, its limited size may not capture the full variability of liver morphology and disease. The LiTS \cite{bilic2023liver} dataset is designed to further facilitate the development and evaluation of liver and liver tumor segmentation algorithms. It comprises 201 abdominal CT scans, including primary and metastatic liver cancers. The training set contains 131 cases with liver and tumor annotations, while the remaining 70 test cases have undisclosed labels. To provide multi-modal imaging for diverse liver lesions, the LLD-MMRI \cite{lou2025sdr} dataset includes 498 patients with diverse liver lesions, each represented by eight MRI scans. Compared with these datasets, the LiMT dataset proposed in this paper offers a similar range of lesion counts and voxel variations as LiTS (Fig. \ref{fig1}), while providing more detailed lesion classification and supporting both benign and malignant lesion analysis. 

In addition, other datasets provide multi-organ annotations. The Medical Segmentation Decathlon (MSD) \cite{simpson2019large} is a multi-modal (MRI, CT) dataset that integrates ten segmentation tasks from different organ datasets, including LiTS \cite{bilic2023liver}. Unlike datasets focusing on pathological abnormalities, CHAOS \cite{kavur2021chaos} provides liver cases without lesions, primarily to support surgical planning. It includes abdominal CT and MR scans from healthy subjects collected at multiple centers for single- or multi-organ segmentation, covering the liver, kidneys, and spleen. A total of 80 cases are included, with 40 of them being CT scan data labeled annotated only for the liver and 40 being annotated MR data with four abdominal organs. However, a learning model trained solely on healthy cases cannot effectively assist doctors in diagnosing diseases. Compared with the CT dataset of CHAOS, which only contains liver annotation, CT-ORG \cite{rister2020ct} provides a CT scan dataset with segmentation annotation of 140 cases of multi-organ, addressing the gap in existing multi-organ CT datasets. In addition, the FLARE dataset \cite{ma2023unleashing} provides 2,300 3D CT images from over 20 centers, 50 labeled images and 2,000 unlabeled images are used for training, and another 50 images are used for validation, which can be used to train semi-supervised models. Furthermore, Luo \textit{et al.} \cite{luo2021word} established a large-scale whole abdominal organ dataset (WORD) to further enhance the development of algorithms and clinical applications for small organs and complex organs. This collection has 150 abdominal CT cases (30,495 images), each annotated for 16 organs, making it arguably one of the most comprehensive abdominal organ labeling datasets available. However, these multi-organ datasets are based on the whole abdominal area, limiting applicability for diagnosing diseases in specific organs.

\subsubsection{Detection task}
The detection task aims to automatically identify anatomical structures and abnormal regions of interest in medical images. As summarized in Table \ref{tab3}, the Anatomy3 dataset \cite{jimenez2016cloud} was the first introduced for such tasks. It provides landmark annotations indicating the locations of various human structures, supporting applications such as image registration and pathological analysis. In addition, the pixel-level annotations in the WORD dataset \cite{luo2021word} can support detection methods by enabling clinicians to quickly and accurately localize organs.

\begin{table}[!t]
	\centering
	\caption{Datasets used for detection}
	\label{tab3}
	\resizebox{0.5\textwidth}{!}{
		\begin{tabular}{lcc}
			\toprule
			Dataset & Annotation type & Class labels \\
			\midrule
			Anatomy3 \cite{jimenez2016cloud} & Landmark & - \\
			\midrule
			DeepLesion \cite{yan2018deeplesion} & Bookmark & \makecell{Lung nodules, Liver lesions, \\ Enlarged lymph nodes, \\ Kidney lesions, Bone lesions, \\ and so on} \\
			\midrule
			WORD \cite{luo2021word} & \multirow{3}{*}{Pixel-wise label} & \makecell{Liver, Spleen, Kidney Left, \\  Kidney Right, Stomach, and so on} \\ \cline{3-3}
			LiTS \cite{bilic2023liver} &  & Tumor, Liver \\ \cline{3-3}
			LiMT &  & \makecell{Liver, Hepatocellular carcinoma, \\ Hepatic metastases, Hepatic cyst, \\ Cavernous hemangioma of liver} \\ 		 
			\bottomrule
	\end{tabular}}
\end{table}

However, lesion detection is more directly relevant to clinical practice. The DeepLesion dataset \cite{yan2018deeplesion} provides multi-class, lesion-level annotations and is designed to develop a large-scale, unified framework for general lesion detection. It enables more accurate and automated measurement of lesion sizes throughout the body, facilitating preliminary whole-body cancer assessment. This dataset contains 32,120 CT slices from 4,427 patients, with a total of 32,735 lesion regions annotated in the form of bookmarks, containing multiple lesion types such as lung nodules, liver tumors, enlarged lymph nodes, etc. It is worth noting that the lesion labels provided by this dataset do not represent an exhaustive list of all lesion annotations in the image. Clinicians usually omit annotations of parts of lesions in order to simplify follow-up research for lesion matching and growth tracking. Therefore, the continued collection of lesions omitted by clinicians remains an important focus for future work.

\subsubsection{Classification task}
Tumor classification tasks can provide tumor type information to assist diagnosis \cite{han2022effective,hao2024ssdc,xing2025re,chen2025haif,zhang2025pure,yuan2024sd,cao2025nerf,yuan2025msp}. As shown in Table \ref{tab4}, Chen \textit{et al.} \cite{chen2019feature} collected 190 portal-phase CT scans containing three lesion types and normal cases. Hussain \textit{et al.} \cite{hussain2022computer} compiled 1,000 CT images from 100 patients, covering two benign and two malignant tumor types with support from the Radiology Department of Nishtar Medical University, Pakistan. However, the dataset does not specify the lesion-type distribution, which may result in class imbalance. In addition, this dataset includes only 10 CT scans per patient for liver tumors, limiting contextual information and posing challenges for real clinical applications. Naeem \textit{et al.} \cite{naeem2020machine} developed an MR–CT fusion dataset with support from the Radiology Department of Bahawal Victoria Hospital, Pakistan, including three benign and three malignant liver tumor types. For each lesion type, 100 MR and 100 CT images were collected. However, MR–CT pairing requires manual intervention, and some patients cannot undergo MR scanning, limiting the dataset’s practical applicability.

\begin{table*}[!t]
	\centering
	\caption{Datasets used for classification}
	\label{tab4}
	\setlength{\tabcolsep}{3pt}
	\begin{tabular}{lcccc}
		\toprule
		Dataset & Type of lesion & Number of each type & Data format & Public/Private \\
		\midrule
		Chen \textit{et al.} \cite{chen2019feature} & Metastases, Hemangiomas,  Hepatocellular carcinoma, Healthy & \makecell{35 scans, 40 scans, 62 \\ scans, 53 scans} & 3D & Private\\
		\midrule
		Naeem \textit{et al.} \cite{naeem2020machine} & \makecell{hepatocellular adenoma,  hemangioma, cyst,\\ hepatocellular carcinoma, hepatoblastoma, metastasis} & 200 slices & 2D & Private\\
		\midrule
		Hussain \textit{et al.} \cite{hussain2022computer} & \makecell{Primary liver malignancies, Metastatic Hepatic\\ adenocarcinomas, Hepatic cysts, Hemangiomas} & - & 2D & Private \\
		\midrule
		LiMT (Ours) & \makecell{Hepatocellular carcinoma, Hepatic metastases ,  \\Hepatic cyst, Cavernous hemangioma of liver, Normal} & \makecell{31 scans, 20 scans,  \\49 scans, 37 scans, 50 scans} & 3D & Public\\
		\bottomrule
	\end{tabular}
\end{table*}

No publicly available liver-related dataset currently supports multiple tasks such as segmentation, classification, and detection. In contrast, we introduce a 3D CT multi-task dataset of 150 cases annotated for both anatomy and lesion types.

\subsection{Liver-related task methods}\label{sec2.2}
In recent years, numerous datasets have supported the development of liver CAD systems. Based on task type, existing liver CAD methods can be broadly categorized into five classes: liver segmentation, liver tumor segmentation, liver and tumor segmentation, liver tumor detection, and liver tumor classification.

Segmentation tasks help clinicians visualize liver shape and structure, facilitating the identification of potential abnormalities. For instance, Kavur \textit{et al.} \cite{kavur2022basic} combined four deep learning models for liver segmentation and evaluated them on two public datasets. Shi \textit{et al.} \cite{shi2021multi} further observed that large tumors introduce significant variability in liver shape, which can hinder model training. To improve the robustness of liver segmentation models in the presence of tumors, a multi-atlas segmentation (MAS) framework based on low-rank tensor decomposition (LRTD) was proposed and evaluated on the SLIVER07 \cite{heimann2009comparison}, 3D-IRCADb \cite{soler20103d}, and LiTS \cite{bilic2023liver} datasets. Zheng \textit{et al.} \cite{zheng20233d} developed a 3D liver segmentation method incorporating multiscale feature fusion and grid attention to address the similar gray values of the liver and surrounding organs, achieving strong performance on both 3D-IRCADb \cite{soler20103d} and SLIVER07 \cite{heimann2009comparison}. To reduce dependence on annotated data, semi-supervised approaches \cite{han2022effective,lyu2025bidirectional} have been proposed that generate pseudo-labels to guide model training. 

Early lesions can induce subtle liver deformations that segmentation algorithms may interpret as noise or insignificant structures, potentially affecting clinical judgment. Consequently, liver tumor segmentation helps physicians assess lesion type, size, and distribution, providing guidance for treatment decisions. Wang \textit{et al.} \cite{wang2024sbcnet} proposed SBCNet, a dual-branch liver tumor segmentation model, demonstrating its effectiveness in handling tumor variability on the LiTS dataset. Di \textit{et al.} \cite{di2022td} developed TD-Net to enhance liver tumor feature representation, achieving improved segmentation accuracy on the LiMT and 3D-IRCADb datasets. SBCNet \cite{wang2024sbcnet} is a dual-branch network that incorporates scale attention and boundary context attention to enhance multi-scale feature extraction and boundary modeling for liver tumor segmentation. It achieves superior performance on the LiTS dataset, notably improving segmentation accuracy for small tumors and low-contrast regions. Similar to SBCNet, which targets small lesions, PCNet  \cite{wei2024boundary} enhances liver tumor segmentation by improving boundary awareness and fine-scale feature extraction, proving particularly effective for low-contrast and small-lesion cases. However, focusing solely on tumor segmentation does not capture their spatial relationship with the liver. In fact, the positional characteristics of tumors within the liver can provide valuable auxiliary information for clinicians, aiding in tumor classification and the assessment of liver tumor invasiveness. The publication of the 3D-IRCADb dataset \cite{soler20103d} in 2012 and the LiTS dataset \cite{bilic2023liver} in 2017 extended liver segmentation annotations to include tumor segmentation annotations. Xi \textit{et al.} \cite{xi2020cascade} introduced a cascaded U-shaped convolutional neural network (U-ResNet), trained on the LiTS dataset \cite{bilic2023liver}, capable of liver and lesion segmentation simultaneously. In addition, on the 3D-IRCADb dataset \cite{soler20103d} and clinical trials, Christ \textit{et al.} \cite{christ2016automatic} and Liu \textit{et al.} \cite{liu2020multi} also validated the proposed method's effectiveness in segmenting the liver and lesions. More recently, the Three-Direction Fusion (TDF) method \cite{zhan2023three} has been proposed, integrating axial, coronal, and sagittal features to enhance segmentation accuracy, offering an efficient multi-view fusion strategy for medical image segmentation.

Unlike the full-pixel segmentation annotations mentioned above, to address the issue of data scarcity, the DeepLesion dataset \cite{yan2018deeplesion} labels lesions using bookmark annotations. It is worth noting that this dataset has its own limitations: not all lesions on each slice are annotated. This limitation may result in incorrect training signals and reduced detection accuracy. Based on the proposed dataset \cite{yan2018deeplesion}, Yan \textit{et al.} \cite{yan2020learning} introduced a multitask approach to effectively learn from multiple heterogeneous lesion datasets to make up for the missing data. Alkhaleefah \textit{et al.} \cite{alkhaleefah2021faster} designed a lightweight architecture that improves average accuracy while reducing network size. Erickson \textit{et al.} \cite{erickson2022class} further addressed the class imbalance problem and improved the recall by performing three experiments. Moreover, the results of detection methods are often presented as the smallest rectangular bounding box of the lesion, which cannot provide pixel-level annotation of the lesion. Although it can promote follow-up research on lesion matching and growth tracking, it shares the same limitations as segmentation methods because the dataset does not provide category labels. That is, it cannot solve the qualitative characterization of lesions and support diagnostic decision-making.

Because early liver lesions often exhibit similar morphology in medical images, while segmentation tasks typically focus on pixel-level details, often neglecting high-level semantic information, making it challenging to distinguish between benign and malignant lesions. Therefore, for the purpose of early diagnosis, the liver lesion classification task can be used as a further development of the above segmentation task. The classification task focuses on understanding the image and helps classify the lesion categories. For instance, a recent study, Hussain \textit{et al.} \cite{hussain2022computer} realized the differentiation of benign and malignant tumors on the collected images through preprocessing steps such as contrast enhancement and correlation-based feature selection (CFS). Similarly, Naeem \textit{et al.} \cite{naeem2020machine} demonstrated the effectiveness of machine learning (ML) methods in liver cancer classification on private datasets using a combined dataset of two-dimensional (2D) CT scans and MRI images. To alleviate the overfitting problem caused by missing data, Chen \textit{et al.} \cite{chen2019feature} utilized a feature fusion adversarial learning network to classify liver lesions based on a private dataset. 

However, a gold-standard medical image dataset is essential for the creation of a reliable CAD system \cite{kadhim2022deep}. Unfortunately, none of the datasets for the above classification methods have been published because of ethical regulations like the Health Insurance Portability and Accountability Act (HIPAA) \cite{ruikar20215k+}. This has hindered the development of CAD by preventing external researchers from accessing datasets to replicate experimental results. At the same time, fair comparisons of algorithms on the same dataset represent another critical factor influencing algorithm development. In such circumstances, this paper presents a publicly available dataset of 3D CT liver images with gold standards (annotations of both anatomy and lesion types). Moreover, unlike the dataset used by Chen \textit{et al.} \cite{chen2019feature}, which only contains the lesion area, our dataset includes complete abdominal CT volumes, aligning better with practical clinical requirements.

\section{Materials and Methods}\label{sec3}
\subsection{Data source}
The study-related data were reviewed and approved by the Institutional Review Board (IRB) of the Research Ethics Committee at the Affiliated Hospital of Jiangsu University, to ensure full compliance with ethical standards for research involving human subjects (Approval number: KY2022K0408). To collect CT cases, the hospital utilized Revolution CT, LightSpeed VCT, and SOMATOM Definition AS+ CT scanners. Table \ref{tab5}, Table \ref{tab6} and Table \ref{tab7} outline the acquisition parameters. The volumetric data for each CT case are stored in DICOM format, and the acquisition parameters are stored in the metadata section of each DICOM file.

\begin{table}[!t]
	\centering
	\caption{CT machine specification}
	\label{tab5}
	\resizebox{0.5\textwidth}{!}{
		\begin{tabular}{cccc}
			\toprule
			Manufacturer & Manufacturer Model & Software Version(s) & Sample Number\\
			\midrule
			\multirow{6}{*}{GE MEDICAL SYSTEMS} & LightSpeed VCT  & 07MW18.4  &5 \\
			\cline{2-3}
			& \multirow{5}{*}{Revolution CT}  &  revo\_ct\_22bc.50 & 42\\
			&  &  revo\_ct\_21b.32     & 6 \\
			&  &  revo\_1.5\_m3c.120   & 11\\
			&  &  revo\_1.5\_m3b.xt.53 & 12\\
			&  &  revo\_1.5\_m3a.46    & 23\\
			\midrule
			SIEMENS & \makecell{SOMATOM Definition \\ AS+}  & syngo CT VA48A &51\\
			\bottomrule
	\end{tabular}}
\end{table}

\begin{table}
	\centering
	\caption{CT acquisition parameters}
	\label{tab6}
	\begin{threeparttable}
		\resizebox{0.5\textwidth}{!}{
			\begin{tabular}{ccccc}
				\toprule
				Manufacturer & Manufacturer Model & ST(mm) & SR(mm) & TR(ms) \\
				\midrule
				\multirow{2}{*}{\makecell{GE MEDICAL \\SYSTEMS}}
				 & LightSpeed VCT & 5 & 0.3 & 40 \\
				\cline{2-5}
				&Revolution CT  &  1/1.25/1.5/5     & 0.23 &28\\
				\midrule
				SIEMENS & \makecell{SOMATOM Definition \\AS+} & 0.6 & 0.3 & 166 \\
				\bottomrule
		\end{tabular}}
		\begin{tablenotes}
			\footnotesize
			\item[*] ST = Slice Thickness (in mm); SR = Spatial resolution (in mm); \\TR = Time resolution (in ms).
		\end{tablenotes}
	\end{threeparttable}
\end{table}

\begin{table}
	\centering
	\caption{Parameters for Contrast-Enhanced Scan Phases}
	\label{tab7}
	\begin{threeparttable}
		\resizebox{0.5\textwidth}{!}{
			\begin{tabular}{ccccc}
				\toprule
				Contrast Injection Rate & Arterial Phase & Portal Venous Phase & Delayed Phase & Contrast Agent Name \\
				\midrule
				\makecell{2.5 ml/s} & 35 s & 80 s & 140 s & Iohexol  \\
				\bottomrule
		\end{tabular}}
	\end{threeparttable}
\end{table}

\subsection{Inclusion criteria}

In this study, several issues may arise during the CT scanning process. For example, metal objects within the body may generate noticeable artifacts in CT images. In addition, considering that training CAD models requires a large volume of specific data, the CT images included in the dataset should ensure reliability, and the lesion types should meet the requirements of clinical diagnosis. As a result, cases included in the dataset must satisfy specific inclusion criteria, categorized into two levels: 1) CT inclusion criteria and 2) liver lesions inclusion criteria.

\subsubsection{CT inclusion criteria}\label{3.2.1}
Unlike most public liver CT datasets that focus on the portal venous phase for liver segmentation and surgical planning, our dataset consists of arterial phase–enhanced CT scans, which emphasize hepatic arterial perfusion and better highlight hypervascular lesions such as hepatocellular carcinoma (HCC). While the arterial phase may present greater challenges for liver or tumor segmentation due to heterogeneous enhancement, it provides superior contrast for detecting and characterizing hypervascular tumors, thereby offering clear advantages in lesion classification and diagnostic accuracy. This distinction underscores the uniqueness and clinical relevance of our dataset. Cysts with a diameter of less than 1 cm may exhibit partial volume effects during CT scanning, which might be misinterpreted as a substantial space-occupying lesion. Excessively thick slices may overlook the characteristics of early lesions. In contrast, thin-slice scanning (3-5 mm or smaller) can more effectively display the CT features of cysts \cite{suetens2017fundamentals}. As a result, all cases in the dataset were scanned with a 1-5 mm slice thickness. During the CT scan process, there are several problems, such as metal implants in the patient's body, improper control of the scanning beam hardness, and patients' inability to control their breathing during the scan. Therefore, cases with significant noise and artifacts were excluded from the dataset.

\begin{figure}
	\centerline{\includegraphics[width=0.9\columnwidth]{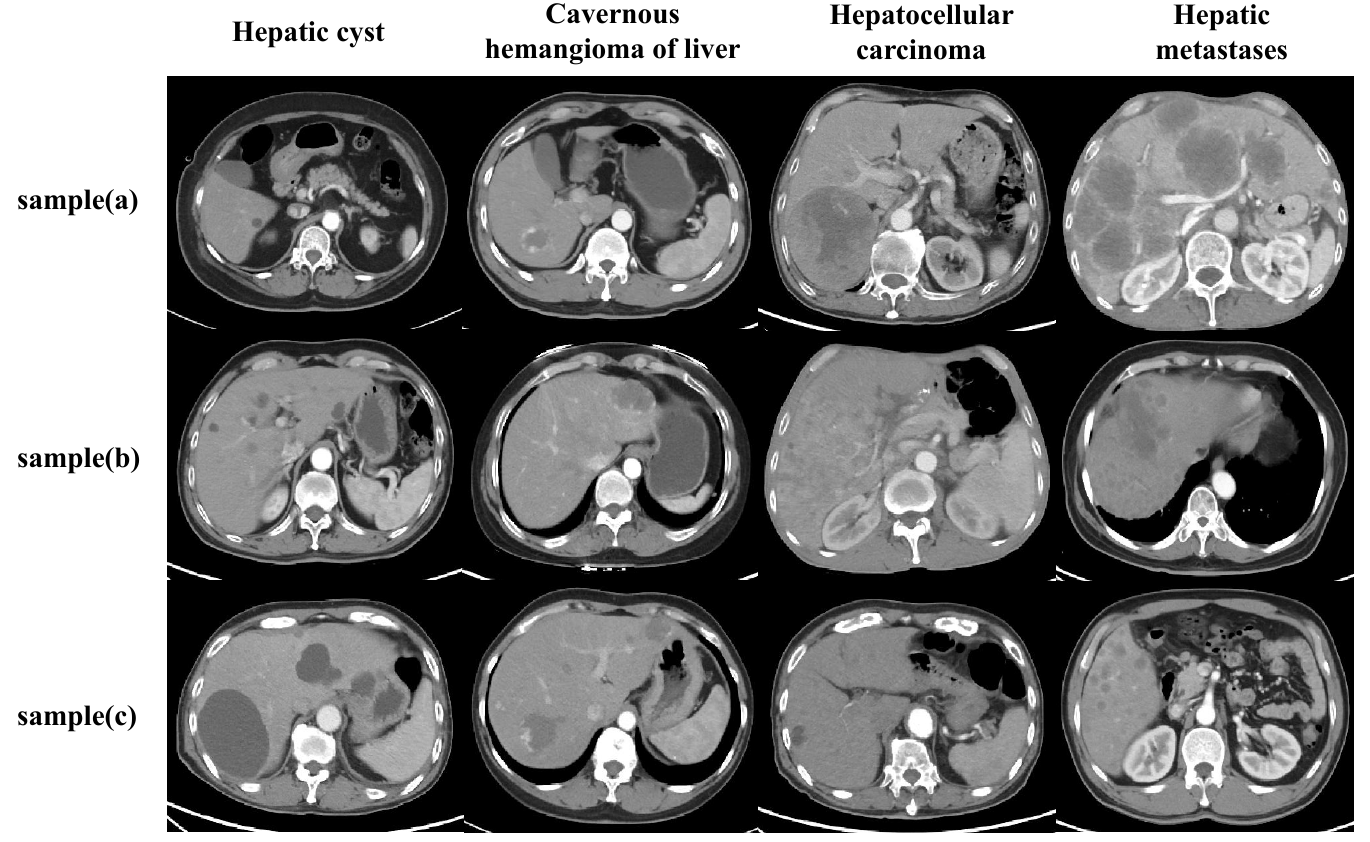}}
	\caption{Samples of four liver lesions in the dataset.}
	\label{fig2}
\end{figure}

\subsubsection{Liver lesions inclusion criteria}
Four types of liver disease are selected for inclusion in the dataset: cavernous hemangioma, hepatocellular carcinoma, hepatic cyst, and hepatic metastases. Fig. \ref{fig2} shows typical cases of the four liver disease types in the dataset of this paper. Among them, hepatocellular carcinoma is a malignant tumor originating from liver tissue, closely associated with hepatitis B or C and cirrhosis, and can be classified into three types on CT imaging: (1) giant type (nodular diameter $\textgreater$ 5cm, such as hepatocellular carcinoma sample (a) in Fig. \ref{fig2}), (2) nodular type (nodular diameter $\leq$ 5 cm, such as hepatocellular carcinoma sample (c) in Fig. \ref{fig2}), and (3) diffuse type (diffuse small nodules distributed throughout the liver, such as hepatocellular carcinoma sample (b) in Fig. \ref{fig2}). Hepatic metastases, caused by malignant tumors spreading to the liver from other organs, are usually multifocal and show irregular edge enhancement in the arterial phase with occasional central unenhanced necrosis forming a typical “bull’s-eye sign,” as illustrated in Fig. \ref{fig2}(a)–(c). Hepatic cysts are benign liver lesions caused by fluid accumulation in the liver, are more common in people aged 30 to 50 years, and large cysts can lead to liver enlargement. As shown in hepatic cyst samples (a) to (c) of Fig. \ref{fig2}, a circular low-density area inside the hepatic parenchyma can be seen, with uniform intracapsular density and visible well-defined boundaries during the arterial phase of an enhanced CT scan. Cavernous hemangioma is a common benign liver tumor, most commonly occurring in women between 30-60 years old. A contrast-enhanced scan is key for CT detection of cavernous hemangioma. As shown in Fig. \ref{fig2}, the cavernous hemangioma of liver samples (a) to (c) in the arterial enhancement stage show scattered spots and nodular enhanced foci appearing at the borders of the tumors, close to the density of large vessels strengthened in the same layer.

Overall, the collected cases ($\geq$20) for each liver lesion type exhibit variation in number, shape, and distribution. The diversity of liver lesion characteristics aids in training CAD models, enhancing their comprehensive understanding of various liver diseases.

\subsection{Annotations}
In addition to CT scan images, the LiMT dataset includes annotations for both anatomy and lesion types. All classification labels in our study are based on pathological or biopsy results, ensuring the high accuracy of each case.

\begin{figure}
	\centerline{\includegraphics[width=0.9\columnwidth]{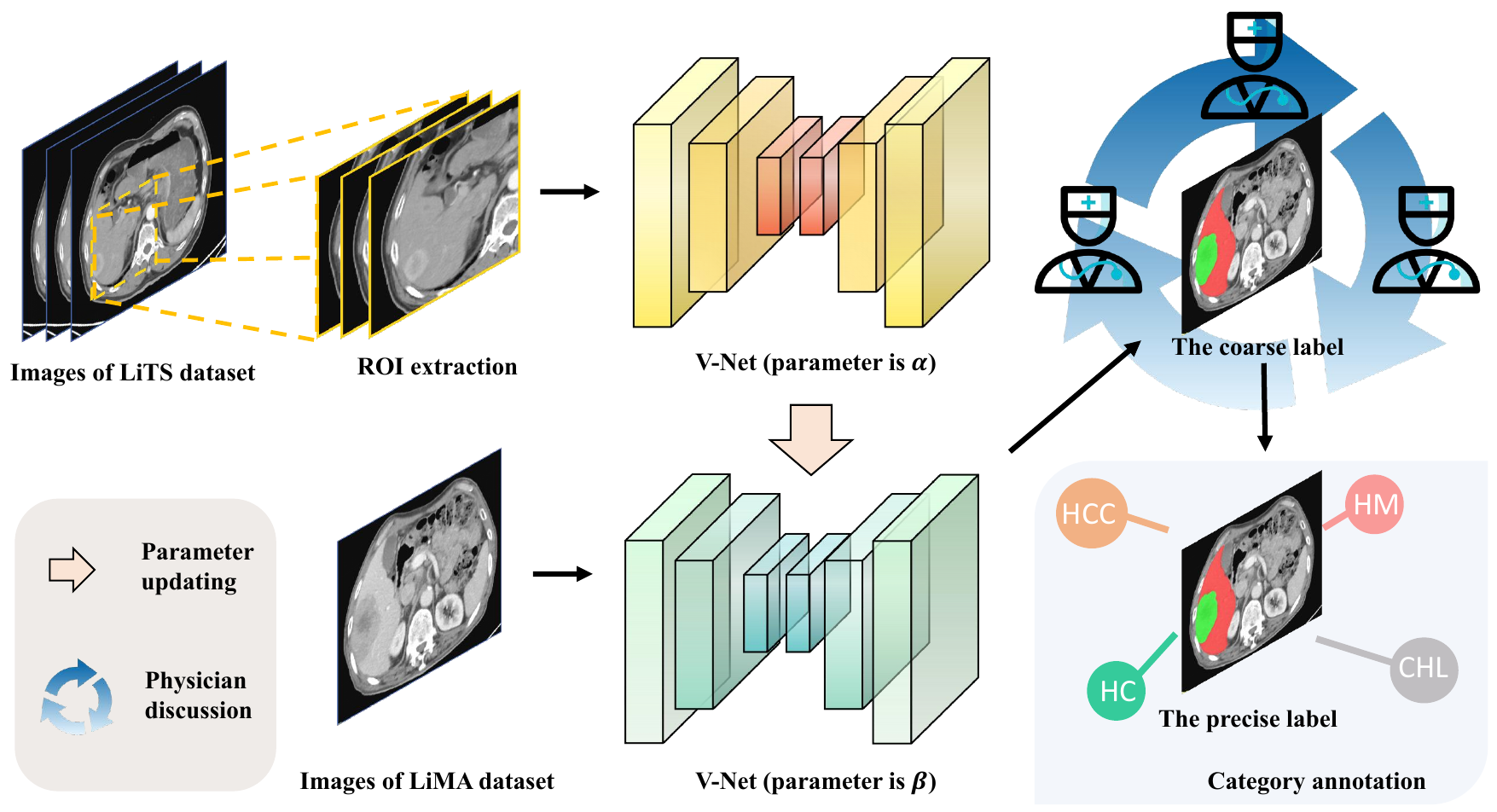}}
	\caption{The flowchart of labeling the LiMT dataset.}
	\label{fig3}
\end{figure}

\begin{figure}
	\centerline{\includegraphics[width=0.9\columnwidth]{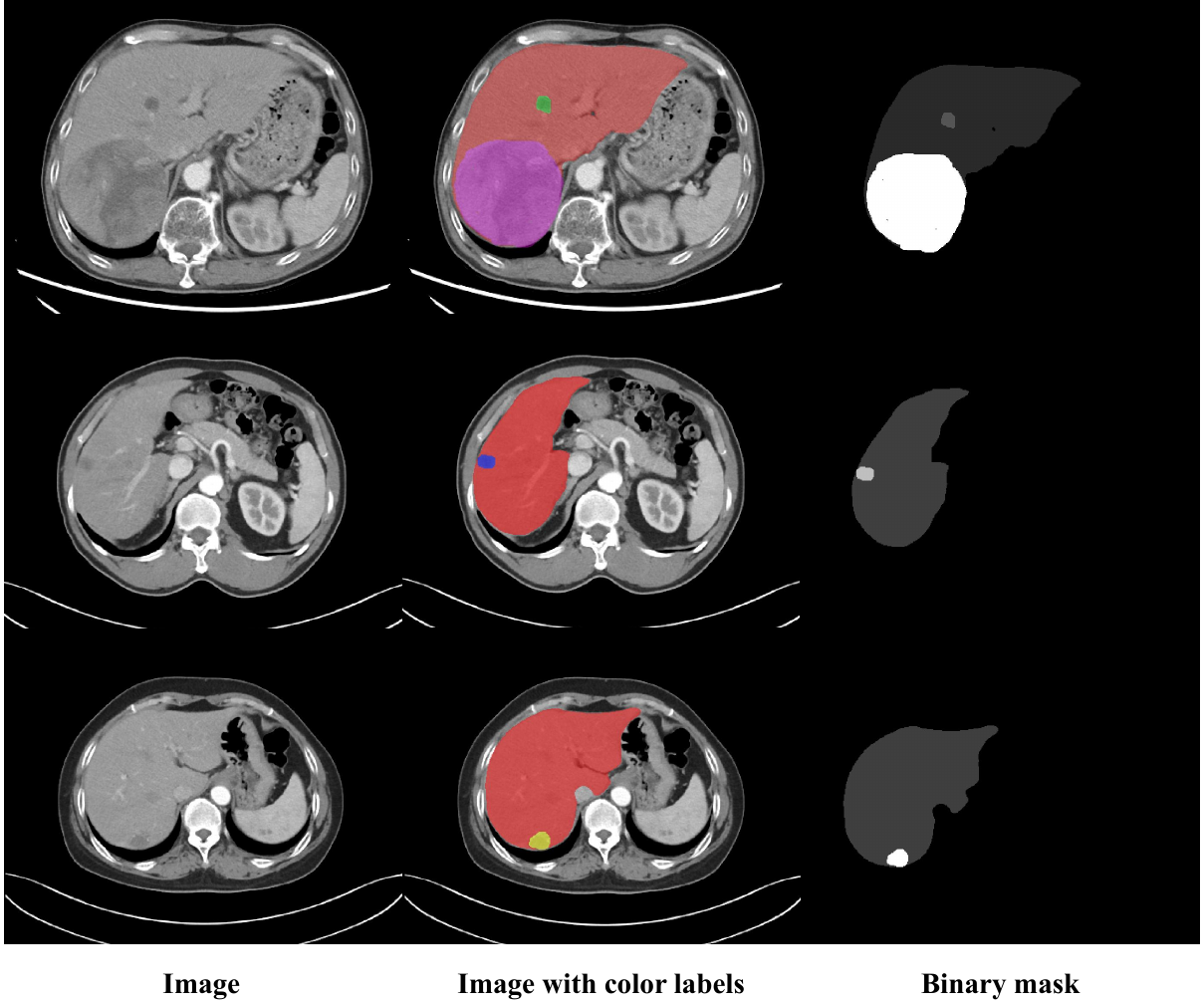}}
	\caption{Images and their corresponding annotations (The liver is annotated in red. Hepatocellular carcinoma, hepatic metastases, hepatic cyst, and cavernous hemangioma of liver are annotated in purple, blue, green, and yellow, respectively.)}
	\label{fig4}
\end{figure}

\begin{figure}
	\centerline{\includegraphics[width=1\columnwidth]{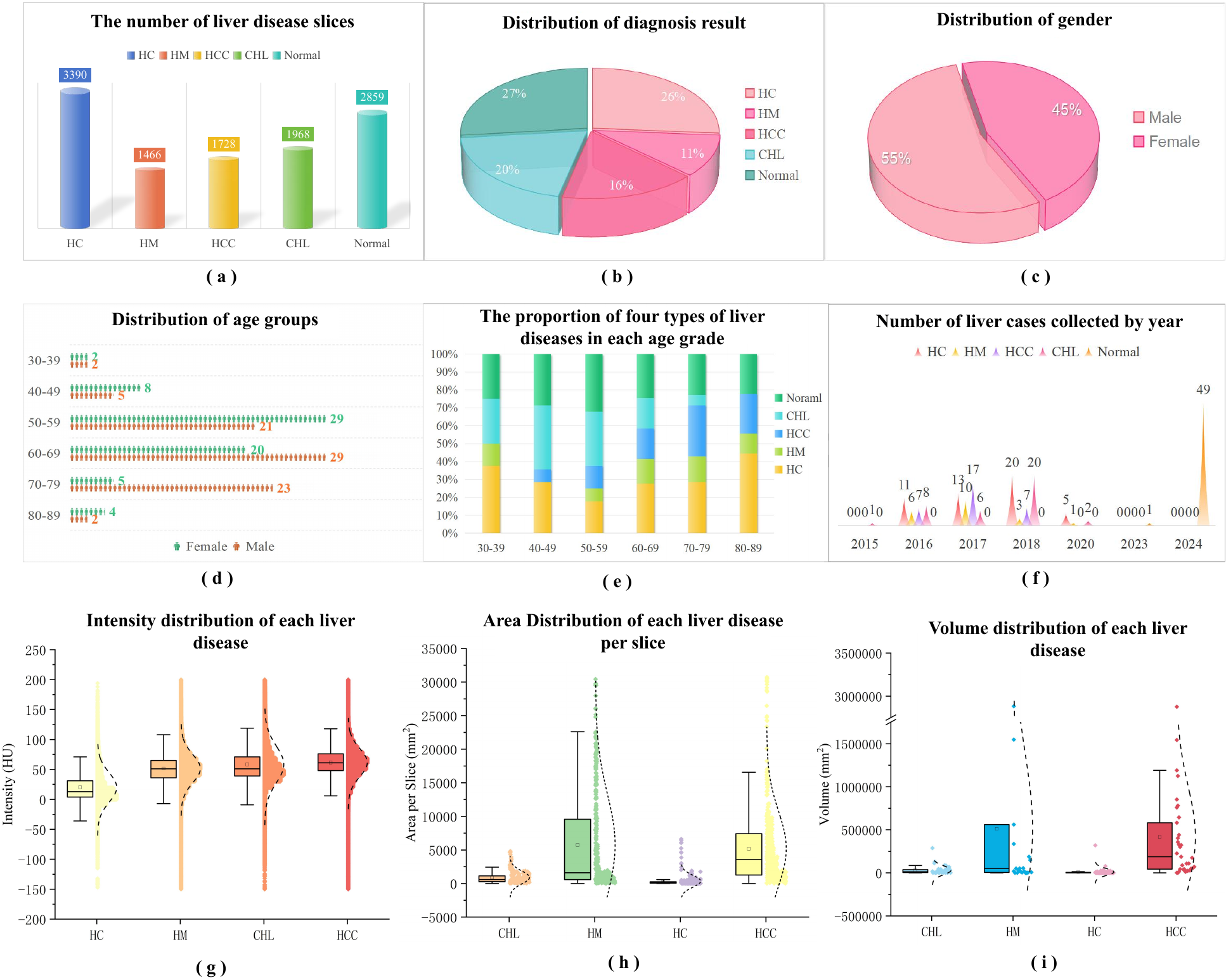}}
	\caption{ Statistical distributions of the dataset. In the figure, hepatocellular carcinoma is denoted as HCC, hepatic metastases as HM, cavernous hemangioma of liver as CHL, and liver cyst as HC. (a) The number of liver disease slices. (b) Distribution of diagnosis results. (c) Distribution of gender. (d) Distribution of age groups. (e) The proportion of four types of liver diseases in each age grade. (f) Number of liver cases collected by year. (g) Intensity distribution of each liver disease. (h) Area distribution of each liver disease in each slice. (i) Volume distribution of each liver disease.}
	\label{fig5}
\end{figure}

The annotation process for segmentation labels involves two steps: a) coarse labeling and b) precise labeling. As shown in Fig. \ref{fig3}, the coarse annotation is first performed using the pretrained segmentation model trained on the LiTS dataset. The pre-annotation code can be found in \footnote{\url{https://github.com/lvchw/LIMT_pre_annotation}}. Specifically, the dataset is preprocessed based on the ground truth provided by the LiTS dataset, and the liver ROI region is extracted to reduce the influence of background noise on model training. Then the preprocessed data is used to train a deep model to learn detailed feature representations. Next, the trained model is used for approximate labeling of our dataset (LiMT). Finally, some mis-segmentation results are removed by the post-processing method \cite{salvi2021impact}. 

Prediction results are not fully accurate, especially for tumor segmentation. Specifically, we present the consistency between the results of deep learning models pre-trained on the LiTS dataset and annotations provided by clinicians: 89.73\% for liver segmentation, whereas tumor segmentation performance is below 10\%. The reason lies in the composition and annotation of the LiTS liver dataset, where only a subset of cases contains tumors, encompassing both primary tumors (e.g., hepatocellular carcinoma, HCC) and secondary tumors (e.g., colorectal cancer metastases). All tumor types are annotated using a single, uniform mask labeled as lesion areas. In contrast, our dataset features multiple tumor categories, each annotated with distinct masks. 

Given these challenges, five diagnostic radiologists participated in the manual correction of the segmentation masks. Two had over 20 years of experience in abdominal imaging, while three had 8–10 years of expertise in liver disease diagnosis and intervention. Before applying majority voting, the inter-rater agreement (Dice Score) among radiologists was 96.15\% for liver regions and 88.46\% for liver lesions, indicating high but not perfect consistency. After majority voting and consensus correction \cite{liao2022learning,liao2024modeling}, the final agreement increased to 98.32\% and 94.78\%, respectively. In cases of disagreement or uncertain boundaries, all annotators jointly reviewed the ambiguous regions to reach a consensus. This process ensured accurate, consistent, and reliable segmentation masks for the dataset. Furthermore, Table \ref{tab8} presents the performance differences of the coarse segmentation algorithm on images with varying slice thicknesses compared with the annotations by clinicians. It can be observed that the algorithm performs significantly better on thinner slices, while its performance declines when the slice thickness increases to 5 mm. This is mainly due to the characteristics of the pretraining dataset (LiTS), which predominantly consists of thin-slice CT scans with a median slice thickness of approximately 0.76 mm and relatively few thick-slice samples. Consequently, the model is better adapted to high-resolution inputs. ITK-SNAP software \cite{yushkevich2016itk} is utilized to adjust the approximate labeling. Fig. \ref{fig4} shows the CT images and the annotations corresponding to them.

\begin{table}
	\centering
	\caption{Liver segmentation performance variance of coarse segmentation algorithms across images with different slice thicknesses}
	\label{tab8}
	\begin{threeparttable}
		\resizebox{0.5\textwidth}{!}{
			\begin{tabular}{cccccc}
				\toprule
				\makecell{Thicknesses [mm]} & 0.625 & 1 & 1.25 & 1.5 & 5 \\
				\midrule
				\makecell{Dice [\%]} & 86.38 & 90.35 & 90.61 &88.56& 79.14  \\
				\bottomrule
		\end{tabular}}
	\end{threeparttable}
\end{table}

\subsection{Data anonymization}
The DICOM header file contains patient-related information such as patient ID, name, gender, date of birth, and age. It also contains the image size and various acquisition parameters for the current investigation \cite{varma2012managing}. Therefore, to preserve patient privacy, the CT image cases in the dataset must be anonymized. To achieve this, confidential data related to the scan center is removed by reassigning specific fields when reading the entire DICOM header file.

\section{Dataset} \label{sec4}
\subsection{Statistical information}
The dataset contains 150 anonymized CT cases. Since some cases include multilabel annotations for liver lesions, it includes 49 cases of hepatic cysts, 37 cases of cavernous hemangioma, 31 cases of hepatocellular carcinoma, 20 cases of hepatic metastases and 50 normal cases. In addition to image data, each image is accompanied by precise annotations, provided in two formats: a) color-coded annotations in ITK-SNAP for labeling purposes, and b) binary (black-and-white) annotations in ITK-SNAP for use as main images. Fig. \ref{fig5} presents basic information such as the number of liver lesion slices and the distribution of diagnostic results, as well as key statistical data on the distribution of different lesions.

\begin{table*}[htbp]
	\centering
	\caption{Performance comparison of different segmentation methods on the LiMT dataset for liver and tumor segmentation}
	\label{tab9}
	\begin{threeparttable}
		\resizebox{\textwidth}{!}{
			\begin{tabular}{cc|cccc|cccc}
				\toprule
				\multirow{2}{*}{Category} & \multirow{2}{*}{Methods} & 
				\multicolumn{4}{c|}{\textbf{Liver Segmentation}} & 
				\multicolumn{4}{c}{\textbf{Tumor Segmentation}} \\
				\cmidrule(lr){3-6} \cmidrule(lr){7-10}
				& & Dice[\%] & Jaccard[\%] & ASD[mm] & HD95[mm] 
				& Dice[\%] & Jaccard[\%] & ASD[mm] & HD95[mm] \\
				\midrule
				\multirow{2}{*}{\textbf{Pre-trained Tools}} 
				& nn-UNet \cite{isensee2021nnu} & 89.82 & 83.27 & 1.91 & 13.28 & 54.99 & 42.62 & 8.02 & 40.53 \\
				& TotalSegmentator \cite{wasserthal2023totalsegmentator} & 89.98 & 83.34 & 2.33 & 9.91 & 44.90 & 33.11 & 12.38 & 37.75 \\
				
				\midrule
				\multirow{4}{*}{\textbf{Training Networks}} 
				& 3D U-Net \cite{cciccek20163d} 
				& 96.48$\pm$0.38 & 92.19$\pm$0.40 & 4.48$\pm$1.01 & 1.56$\pm$0.43 
				& 55.91$\pm$5.34 & 38.78$\pm$5.09 & 9.49$\pm$4.54 & 36.37$\pm$10.91 \\
				
				& V-Net \cite{milletari2016v} 
				& 96.43$\pm$0.36 & 91.87$\pm$0.62 & \textbf{3.88$\pm$1.25} & 1.48$\pm$0.50 
				& 54.80$\pm$5.49 & 36.15$\pm$6.55 & 7.08$\pm$4.26 & 33.50$\pm$8.08 \\
				
				& nn-UNet (trained) \cite{isensee2021nnu} 
				&  \textbf{97.03$\pm$0.43} &  \textbf{93.28$\pm$0.55} & 3.93$\pm$1.73 & \textbf{0.43$\pm$0.32} 
				& \textbf{58.80$\pm$4.28} & \textbf{40.73$\pm$4.89} & \textbf{6.57$\pm$3.75} & \textbf{30.85$\pm$7.25} \\
				
				& Swin Transformer \cite{liu2021swin} 
				& 96.64$\pm$0.51 & 92.35$\pm$0.48 & 5.78$\pm$1.51 & 2.01$\pm$1.41 
				& 50.44$\pm$8.49 & 35.83$\pm$9.51 & 9.29$\pm$4.02 & 35.57$\pm$9.74 \\

				\bottomrule
			\end{tabular}
		}
	\end{threeparttable}
\end{table*}

\begin{table}
	\centering
	\caption{Performance comparison of different baseline classification methods on the LiMT dataset}
	\label{tab10}
	\begin{threeparttable}
		\resizebox{0.5\textwidth}{!}{
			\begin{tabular}{ccccc}
				\toprule
				Methods & Accucary[\%] & Precision[\%] & Recall[\%] & F1[\%] \\
				\midrule
				DenseNet \cite{huang2017densely} & 60.55$\pm$3.75 & 52.26$\pm$1.37 & \textbf{55.65$\pm$1.86} &\textbf{52.47$\pm$1.38}\\
				ViT \cite{dosovitskiy2020image} & 58.32$\pm$6.55 & 50.94$\pm$2.53 & 53.39$\pm$3.08 &50.81$\pm$3.31\\
				ResNet \cite{he2016deep} & \textbf{73.66$\pm$4.79} & \textbf{59.23$\pm$8.10} & 53.27$\pm$2.24 & 51.59$\pm$2.81\\
				\bottomrule
		\end{tabular}}
	\end{threeparttable}
\end{table}

\begin{table}
	\centering
	\caption{Performance comparison of different baseline lesion detection methods on the LiMT dataset}
	\label{tab11}
	\begin{threeparttable}
		\resizebox{0.40\textwidth}{!}{
			\begin{tabular}{cccccc}
				\toprule
				Methods & Precision & Recall & MAP50  & Params(M)\\
				\midrule
				Yolov8 \cite{yolov8}  & 0.763 & 0.732 & 0.773& 3.006\\
				YoloV11 \cite{yolov11} & 0.756 & 0.710 & 0.761& \textbf{2.583}\\
				RT-DETR \cite{zhao2024detrs} & 0.738 & 0.663 & 0.697& 31.992\\
				DEIM \cite{huang2025deim}    & \textbf{0.780} & \textbf{0.915} & \textbf{0.791}& 3.725\\
				\bottomrule
		\end{tabular}}
	\end{threeparttable}
\end{table}

\begin{table}[t]
	\centering
	\caption{Performance comparison between single-task and multi-task models on segmentation and classification tasks.}
	\label{tab12}
	\resizebox{\columnwidth}{!}{%
		\begin{tabular}{lcccc}
			\toprule
			\multicolumn{5}{c}{\textbf{Segmentation Performance}} \\
			\midrule
			\textbf{Method} & Dice [\%] & Jaccard [\%] & ASD [mm] & HD95 [mm] \\
			\midrule
			Segmentation-only \cite{isensee2021nnu} & 58.80$\pm$4.28 & 40.73$\pm$4.89 & 6.57$\pm$3.75 & 30.85$\pm$7.25 \\
			\textbf{Multi-task \cite{chen2023multi}} & \textbf{60.04$\pm$4.18} & \textbf{43.58$\pm$4.52} & \textbf{5.95$\pm$4.02} & \textbf{28.73$\pm$6.98} \\
			
			\midrule
			\multicolumn{5}{c}{\textbf{Classification Performance}} \\
			\midrule
			\textbf{Method} & Accuracy [\%] & Precision [\%] & Recall [\%] & F1 [\%] \\
			\midrule
			Classification-only \cite{he2016deep} & 73.66$\pm$4.79 & 59.23$\pm$8.10 & \textbf{53.27$\pm$2.24} & 51.59$\pm$2.81\\
			\textbf{Multi-task \cite{chen2023multi}} & \textbf{75.83$\pm$4.05} & \textbf{60.91$\pm$7.58} & 53.03$\pm$2.35 & \textbf{53.40$\pm$2.32} \\
			\bottomrule
		\end{tabular}%
	}
\end{table}

\subsection{Benchmark}
\subsubsection{Experimental Details and Metrics}
All experiments were conducted on an NVIDIA RTX A6000 GPU equipped with 48 GB of memory. For the segmentation task, models were trained on volumetric inputs with a patch size of $96\times96\times96$. All images were cropped according to liver masks to remove irrelevant background regions, and random patches were sampled during training to enhance data diversity and model robustness. Inference was performed using a sliding-window approach, and final predictions were obtained by aggregating patch-wise results. For the classification task, 3D volumetric inputs were center-cropped and intensity-normalized before being fed into the network. Random flipping and rotation were applied for data augmentation. The AdamW optimizer was adopted with an initial learning rate of 0.0001, a batch size of 4, and a cosine learning rate schedule. Training was performed for 2,000 epochs under deterministic settings to ensure reproducibility. For the detection task, the AdamW optimizer was adopted for all models. Specifically, YOLOv8, YOLOv11, and RT-DETR were trained with a batch size of 32 and a learning rate of 0.01, while DEIM used a batch size of 8, an initial learning rate of $8\times10^{-4}$, and a weight decay of $1\times10^{-4}$. For all detection models, the AdamW optimizer was employed with task-specific hyperparameters, and training was performed under deterministic settings to ensure reproducibility.

For segmentation tasks, performance was evaluated using multiple metrics: Dice, Jaccard, ASD, and HD95. Dice and Jaccard quantify the overlap between predicted and ground truth regions, while ASD measures the average surface distance between them. HD95 focuses on the 95th percentile of the Hausdorff distance, which highlights the maximum discrepancy and provides insight into worst-case segmentation performance. For multi-label classification tasks, accuracy, precision, recall, and F1-score were employed. Accuracy represents the proportion of correctly predicted labels, precision indicates the proportion of true positives among predicted positives, recall assesses the proportion of actual positives correctly identified, and F1-score combines precision and recall into a single metric. More detailed benchmark methods, results, and experimental parameters are available in our benchmark code at \footnote{\url{https://github.com/lvchw/LIMT_benchmark}}.

\subsubsection{Results and Analysis}

For the segmentation task, a comparison was performed between using pre-trained models (nn-UNet \cite{isensee2021nnu} and TotalSegmentator \cite{wasserthal2023totalsegmentator}) for direct prediction and five-fold cross-validated training of models (3D U-Net \cite{cciccek20163d}, V-Net \cite{milletari2016v}, nn-UNet \cite{isensee2021nnu}, and Swin Transformer \cite{liu2021swin}), with the results summarized in Table \ref{tab9}. The trained nn-UNet achieved the best performance in liver segmentation, attaining the highest Dice and Jaccard scores along with superior boundary accuracy. Although tumor segmentation remains challenging, it still outperformed other methods in Dice, Jaccard, and boundary metrics, demonstrating more stable and accurate results. Overall, task-specific trained convolutional networks outperform pre-trained tools and Transformer models, particularly for segmenting complex structures.

\begin{figure}[!t]
	\centering
	\centerline{\includegraphics[width=\linewidth]{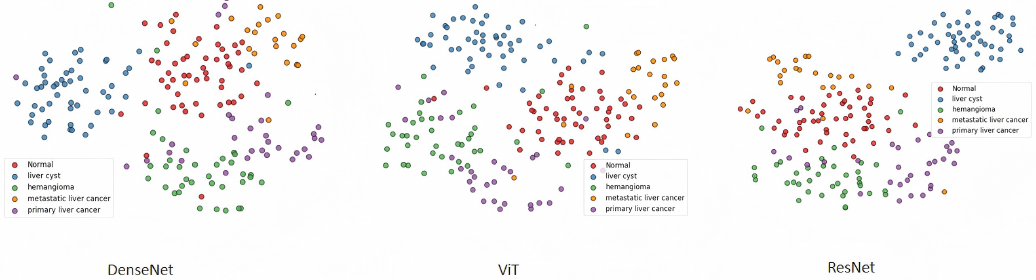}}
	\caption{t-SNE visualization of different classification methods on four diseases and normal cases.}
	\label{fig6}
\end{figure}

For the classification task, we utilized classic classification networks, including DenseNet \cite{huang2017densely}, ViT \cite{dosovitskiy2020image}, and ResNet \cite{he2016deep}. Table \ref{tab10} presents the results of lesion classification. ResNet achieves the highest performance with an average accuracy of 73.66\% and a precision of 59.23\%, indicating reliable overall classification and better discrimination of positive samples. However, the accuracy remains modest, likely due to the inherent difficulty of the task—liver lesions often exhibit small sizes, subtle contrast differences, and high inter-class similarity, which make fine-grained classification challenging even for deep convolutional networks. In contrast, ViT performs poorly, with an accuracy of only 58.32\%, accompanied by a high standard deviation, indicating potential overfitting and insufficient generalization ability. We speculate that ViT underperforms due to overfitting on the limited LiMT dataset and its insufficient ability to capture fine-grained local details crucial for small liver lesion identification. Additionally, qualitative results of the three baseline classification methods are shown in Fig. \ref{fig6}. All methods perform relatively well on liver cysts, likely because they typically appear as round or oval structures with well-defined boundaries and homogeneous internal density in contrast-enhanced CT images. In contrast, their performance on the other three lesion types and normal samples is poorer, which may be attributed to higher inter-class similarity and subtle lesion features.

For the detection task, as shown in Table \ref{tab11}, DEIM achieves the highest precision (0.780), recall (0.915), and MAP50 (0.791), benefiting from its enhanced feature interaction and attention mechanisms that improve detection of small or low-contrast lesions. The YOLO series (YOLOv8 and YOLOv11) maintains competitive precision and MAP50 due to their efficient single-stage detection framework but shows lower recall, indicating some missed lesions. RT-DETR performs worst, particularly in recall and MAP50, likely reflecting the difficulty of transformer-based detectors in capturing small lesions without extensive fine-tuning. Overall, the results demonstrate that attention-enhanced, multi-scale feature extraction methods, such as DEIM, are more effective for lesion detection in the LiMT dataset.

To evaluate the benefits of joint learning, Table \ref{tab12} presents a comparison between single-task and multi-task model (\cite{chen2023multi}) on segmentation and classification tasks. The multi-task model outperformed single-task models in segmentation, achieving higher Dice and Jaccard scores while reducing ASD and HD95, indicating more accurate and precise boundaries. For classification, the multi-task approach improved accuracy, precision, and F1 score compared to the classification-only model, demonstrating more reliable predictions. Overall, jointly learning segmentation and classification benefits both tasks, enhancing performance and stability over single-task models.

\section{Discussion}
In this study, we propose a multi-task dataset, LiMT, for the segmentation and classification of liver lesions. The dataset covers four different types of liver diseases, including hepatocellular carcinoma, hepatic metastasis, hepatic cyst, and cavernous hemangioma, and provides multi-label annotations. Unlike traditional single-task datasets, the LiMT dataset not only includes a diversity of lesion types but also contains annotations of anatomical structures, providing richer multimodal information for liver disease diagnosis. The design of this dataset effectively supports multi-task learning, enhancing the model's generalization and robustness, especially when dealing with complex liver tumor segmentation and multi-label classification tasks.

In the experiments, we evaluated the performance of the LiMT dataset on liver segmentation, lesion classification and detection using several classical baseline models. The results indicate significant performance differences across the various models on different tasks. Particularly in the liver segmentation task, the nn-UNet model showed strong advantages in different metrics. These findings highlight the importance of selecting the appropriate model architecture based on specific task requirements, especially when the dataset contains medical images with high heterogeneity and complex backgrounds.

Although the LiMT dataset provides high-resolution voxel-level annotations and lesion category labels for liver disease segmentation and classification tasks, there are still some limitations. Firstly, the dataset is relatively small and currently comes from a single medical center, which may limit its representativeness, especially in terms of clinical case diversity and population bias. Additionally, the dataset's coverage of certain liver disease types remains incomplete, as it does not include a variety of liver lesions such as simple liver cysts, hepatic adenomas, and pseudotumors. This could restrict the model's performance in real-world clinical applications, particularly when faced with more complex or rare lesions. 

To address these limitations, future work will focus on both technical and data expansion aspects:

\textbf{(1) Technical aspect}

From a technical perspective, future research will focus on generation-based augmentation, semi-supervised learning, self-supervised representation learning, and domain adaptation to enhance model generalization and robustness. From the generation perspective, label-free methods \cite{hu2023label} and diffusion-based models \cite{kazerouni2022diffusion} can generate realistic and diverse liver lesions, enriching dataset variability and compensating for underrepresented disease types. In the semi-supervised learning domain, recent studies \cite{han2024deep,huang2025uncertainty,pan2025dusss} demonstrate that leveraging unlabeled data through uncertainty-aware and vision-language co-supervision can effectively boost segmentation under limited annotations. Meanwhile, self-supervised approaches \cite{gui2024survey,chen2024collaborative,perez2025exploring} show that large-scale pretraining with masked modeling, multimodal alignment, and text-guided supervision can learn generalizable and semantically rich feature representations from unlabeled medical data, thereby improving adaptability to diverse liver pathologies. Besides, although the inclusion of data from multiple scanners and varying slice thicknesses enhances the clinical representativeness of LiMT, it also introduces intensity and resolution inconsistencies that may affect model convergence and generalization. Future work will incorporate domain adaptation and slice normalization strategies \cite{li2024comprehensive} to mitigate these effects and further strengthen cross-center robustness. Integrating these strategies can alleviate data scarcity, reduce annotation dependency, and enhance the generalization of liver disease analysis models.

\textbf{(2) Data aspect}
\begin{itemize}
	\item 	\textbf{Integrating multi-center data:} We are collaborating with multiple medical centers to expand the dataset by integrating diverse clinical cases to address the limitations of single-center studies, such as population bias. This will enhance the dataset's representativeness of real-world clinical conditions and support the development of more generalizable and robust deep learning models.
	\item  \textbf{Covering a wider range of diseases:} To enhance the dataset's diversity and clinical relevance, we aim to include a broader range of liver diseases, such as simple liver cysts, focal nodular hyperplasia, hepatic adenoma, adenomatosis, pseudotumors, and other hepatic conditions. This will better capture the heterogeneity of liver diseases encountered in clinical practice, enhancing the model's capability to tackle diverse diagnostic challenges and identify subtle variations in liver pathology.
	\item  \textbf{Acquiring finer and more comprehensive data:} We aim to build a finer-grained liver CT dataset including thin-slice, multi-phase enhanced, and separately curated non-contrast scans. Thin slices improve lesion and vessel annotation, multi-phase scans reveal dynamic vascular patterns and disease progression, and plain CT offers a cost-effective baseline highlighting density changes, calcifications, and cystic lesions, supporting robust detection and wider clinical research.
	\item  \textbf{Iteratively optimizing lesion annotations} For the current dataset, despite significant efforts, false negatives may still be present in the segmentation labels due to the following reasons: (1) Some lesions are extremely small or located near blood vessels, which makes accurate annotation challenging; (2) Certain atypical lesions are difficult to discern with the naked eye. We plan to refine annotations in the future by manually correcting labels with the aid of multi-phase enhanced scan information and by introducing automated methods for noisy label correction to further improve the quality of the segmentation labels.
\end{itemize}

\section{Conclusion} \label{sec5}
In this work, we introduce LiMT, a large-scale multi-task dataset specifically curated to advance liver computer-aided diagnosis (CAD). LiMT encompasses multiple complementary tasks, including liver and tumor segmentation, lesion detection, and multi-label lesion classification. It facilitates both single-task development and cross-task correlation studies while mitigating dataset heterogeneity. Each LiMT case contains detailed clinical information with carefully validated annotations, and the dataset is expected to serve as a valuable reference for medical imaging research and education. Future work will focus on expanding the LiMT dataset through multi-center collaboration, larger and more diverse clinical cases, and high-quality multi-phase imaging, while integrating advanced generative, semi-supervised, and self-supervised learning techniques. These efforts aim to enhance the dataset’s diversity and completeness, reduce annotation dependence, and improve model robustness and generalization for real-world liver disease analysis.

\section*{Acknowledgment}
The authors would like to thank the clinicians from the Department of Radiology at the Affiliated Hospital of Jiangsu University.

\section*{References}
\bibliographystyle{IEEEtran}
\bibliography{refs}

\end{document}